\documentclass{svproc}

\usepackage{url}

\usepackage{amsmath,graphicx}
\usepackage{color}
\usepackage[english]{babel}
\usepackage[utf8x]{inputenc}
\usepackage[T1]{fontenc}
\usepackage{subcaption}
\usepackage{tabu} 
\usepackage[a4paper,top=3cm,bottom=2cm,left=3cm,right=3cm,marginparwidth=1.75cm]{geometry}
\usepackage{amsmath}
\usepackage{url}

\usepackage[colorinlistoftodos]{todonotes}
\usepackage[colorlinks=true, allcolors=blue]{hyperref}

\mainmatter              

\begin{document}
\title{Detection of distal radius fractures trained by a small set of X-ray images and Faster R-CNN.}
\titlerunning{Detection of distal radius fractures} 
\author{Erez Yahalomi \inst{1} \and Michael Chernofsky \inst{2} \and Michael Werman\inst{1}}
\authorrunning{Erez Yahalomi et al} 
\tocauthor{Erez Yahalom, Michael Chernofsky, and Michael Werman} 
\institute{School of Computer Science and Engineering. The Hebrew university of Jerusalem, Jerusalem, Israel
\and
Hadassah-Hebrew University Medical Center, Jerusalem, Israel}

\maketitle
 
\begin{abstract}

Distal radius fractures are the most common fractures of the upper extremity in humans. As such, they account for a significant portion of the injuries that present to emergency rooms and clinics throughout the world. We trained a Faster R-CNN, a machine vision neural network for object detection, to identify and locate distal radius fractures in anteroposterior X-ray images. We achieved an  accuracy of 96\% in identifying fractures and mean Average Precision, mAP,  of 0.866. This  is significantly more accurate than the  detection achieved by physicians and radiologists.  These results were obtained by training the deep learning network with only 38 original images of anteroposterior hands X-ray images with fractures. This opens the possibility to detect  with this type of neural network rare diseases or rare symptoms of common diseases ,  where only a small set of diagnosed X-ray images could be collected for each disease.
\end{abstract}
\makeatletter
\def\blfootnote{\xdef\@thefnmark{}\@footnotetext}
\makeatother
\blfootnote{This research was supported by the Israel Science Foundation and by the  Israel Ministry of Science and Technology.}

\section{Introduction}
Distal radius fractures at the  wrist are the most common fractures of the upper extremity in humans. As such, they account for a significant portion of the injuries that present to emergency rooms and clinics throughout the world.
Every year tens of millions of people worldwide suffer hand trauma
\cite{Serviste:13}. 
Diagnosis is made on the basis of X-ray imaging, a technology that remains prevalent despite being developed over one hundred years ago. In England alone about ten million visits per year  to accident and emergency centers involve having an X-ray  mostly to check for bone injury, \cite{Feinmann:15}.
A significant portion, up to 30\%,  of the hand wrist X-ray images are incorrectly diagnosed, \cite{ootes:11}, resulting in many people not receiving the medical treatment they need.

Multiple  X-rays of an injured wrist are typically performed, including anteroposterior (AP), lateral, and oblique images, in order to capture pictures of the bones from different points of view, and optimize fracture recognition.

Treatment of distal radius fractures depends first and foremost upon a diagnosis based on the X-rays. Once a diagnosis is made, the specific treatment  depends upon many factors, including the radiographic nature of the fracture pattern. Treatment can include cast immobilization, closed reduction and casting, closed or open reduction and pin fixation, closed or open reduction and external fixation, as well as various techniques of open reduction and internal fixation. The fundamental principles of treatment are restoration of normal alignment and position of the fracture components, and maintenance of this condition until adequate healing has taken place. 

Despite the  exponential growth of technology, X-ray interpretation remains archaic. Throughout the world we rely on medical doctors to look at and accurately interpret these images.  Most doctors who interpret X-rays are general practitioners, family doctors, or general orthopedic doctors, who may have very limited training in wrist X-ray interpretation. The images themselves are inherently difficult to interpret, and this is compounded in the case of distal radius fractures by the fact that the bones of the wrist obscure each other in X-ray images.  Even radiologists, who are medical specialists whose job it is to interpret diagnostic images,   make mistakes. Fractures are missed, proper diagnoses are not made, and patients suffer, with resulting economic and medicolegal consequences.

Machine learning can provide an effective way to automate and optimize diagnosis of distal radius fractures on the basis of X-rays images, minimizing the risk of misdiagnoses and the subsequent personal and economic damages.


In this paper, we demonstrate automatic computerized detection of anteroposterior (AP) distal radius fractures based on a Faster R-CNN neural network \cite{Shaoqing:15}. We trained a Faster R-CNN, a neural network for object detection, to identify and locate distal radius fractures in anteroposterior X-ray images achieving the excellent
accuracy of 96\% of identifying fractures and a mAP of 0.87.

Distal radius fractures are the most common hand fracture. The computerized  fracture detection system we present has higher accuracy in detecting AP distal fractures than the average radiologist and physician.  

Fracture detection by machine vision is a challenging task. In many cases the fracture's size is small and hard to detect. Moreover, the fractures have a wide range of different shapes.  The classification of an image shape as a fracture also depends in its location.  A shape diagnosed as a fracture in  one location is diagnosed as a normal structure of the hand in a different location. To cope with these challenges, we trained a state of the art neural network for object detection, Faster R-CNN, for two tasks: a. classifying if there is a fracture in the distal radius. b. finding the fracture's location.
The advantage of Faster R-CNN is that it can handle high-resolution images. We trained this network with images with a resolution of up to $1600\times 1600$ pixels.  The ability to process high-resolution images enables Faster R-CNN to successfully detect objects in X-ray images, 
which  reveals less details and produces images with a lower signal to noise ratio than MRI or CT. In addition, this enables Faster R-CNN to detect small objects. With its unique network structure, described in section 3, Faster R-CNN can be trained to a high accuracy in detecting objects with a small number of images. 
In this paper, we show that Faster R-CNN can be trained by a small set of X-ray images to produce high accuracy detection results.
This makes Faster R-CNN an excellent tool to detect rare symptoms or rare diseases in X-ray, MRI and CT images where there are only a small number of available X-ray images of these rare medical cases.

\textbf{
\section{Related work}
}
Vijakumar et al. discuss image pre-processing and enhancement techniques of medical images for removing different types of noise from the images such as Gaussian, salt and pepper etc. \cite{Vijaykumar:10}.
Edward et al \cite{edward2015robust}, introduced automated techniques and methods to verify the presence or absence of fractures. To make these images accurate they applied one or more steps of pre-processing to remove the noise. The existing scheme is modified by a better segmentation and edge detection algorithm to improve efficiency. S. Deshmukh et al \cite{nair2002coronary}, conclude that  Canny Edge detection can be used in detecting fractured bones from X-ray images. 
Wu et al detected fractures in pelvic bones with traumatic pelvic injuries, by automated fracture detection from segmented  tomography (CT) images \cite{wu2012fracture}, based on  a hierarchical algorithm, adaptive windowing, boundary tracing, and the wavelet transform. Fracture detection was performed based on the results of prior pelvic bone segmentation via an active shape model (RASM),
Rathode and  Wahid used to train various classifiers, \cite{rathode2015mri}. 

Object detection is a major area in computer vision field. Traditional features are the histogram of oriented gradients,  scale-invariant features, wavelet transforms etc., but the performance of traditional methods fall short of current work based on  deep learning,  \cite{Shaoqing:15,rizhevsky:12,szegedy2015going,he2016deep}.

Convolutional neural networks have been applied to medical images imaged with  different  techniques in recent years. For example, computed tomography (CT) \cite{hoo2016deep}   and X-rays \cite{bar2015chest,shin2016learning,esteva2017dermatologist}. CNN models are successful in many medical imaging problems. H.C Shin et al \cite{shin2016learning}  used CNNs to study specific detection problems, such as thoraco-abdominal lymph node (LN) detection and interstitial lung diesis (ILD)classification. Esteva et al \cite{esteva2017dermatologist}. used  googlenet\cite{szegedy2015going},  for skin cancer classification, reaching a level of accuracy comparable to dermatologists. Dong et al \cite{dong2017learningrathode2015mri},  use CNNs with 16,000 diagnosed X-ray images, to train a classification model to diagnose multi lung diseases such as bronchitis, emphysema and aortosclerosis.
Sa et al used a Faster R-CNN deep detection network to identify osseous landmark points in the lateral lumbar spine in X-ray images, \cite{sa2017intervertebral}.
\\

\section{Background}

\textbf{Faster R-CNN}
is a state of the art detection network.  It has three parts:
1. A convolutional deep neural network for classification and generating a feature map.
2. A regional proposal network, generating region proposals.    
3. A regressor, finding by regression and additional convolutional layers, the precise location of each object and its classification.
The neural network for classification we used is VGG 16. It contains 16 layers including 3x3 convolution layers, 2x2 pooling layers and fully connected layers with over 144 million parameters.  The convolutional feature map can be used to generate rectangular object proposals.  To generate region proposals a small network is slid over the convolutional feature map output. This feature is fed into two fully connected layers, a box regression layer, and a box classification layer.

\textbf{Anchors}, at each  window location, up to 9 anchors with different aspect ratios and scales give region proposals. The RPN keeps anchors that either has the highest intersection over union (IOU) with the ground truth box or anchors that have IOU overlap of at least 70\% with any positive ground truth. A single ground truth element may affect several anchors.

\textbf{The loss function,} Faster R-CNN is optimized for a multi-task loss function. The multi-task loss function \cite{Shaoqing:15}, combines the losses of classification and bounding box regression.
\begin{equation}\label{first_eqn}
	L(\{p_i\},\{t_i\})=\frac{1}{N_{cls}} \sum_i L_{cls}(p_i,p_i^*)+\lambda \frac{1}{N_{reg}} \sum_i p_i^* L_{reg}(t_i,t_i^*)
\end{equation}
$i$ - index of anchors in mini batch,
$p_i$ - predicted probabilities of anchor being an object,
$p_i^*$ - is 1 if anchor is positive and 0 if anchor is negative.   
$t_i$ - is a vector representing the 4 parameterized coordinates of the predicted bounding box, 
$t_i^*$ -   the ground truth vector of the coordinates associated with positive anchors,
$L_{cls}$ - classification log loss over the two classes (object vs.  no object).

\textbf{Regression loss,} the output of the regression determines a predicted bounding box and the regression loss indicates the offset from the true bounding box. The regression loss [3] is;
   
\begin{equation}\label{second_eqn}
L_{reg}(t_i,t_i^*) = R(t_i-t_i^*)
\end{equation}
Where R is the robust loss function (smooth L1) defined in \cite{girshick2015fast}. 

\textbf{Training RPN}, RPN is trained by propagation and stochastic gradient descent (SGD).  New layers are randomly initialized by zero-mean Gaussian distributions. All other layers are initialized using the pre-trained model of image-net classification.

\begin{figure}[h!]
   \centering
  \includegraphics[width=0.6\textwidth]{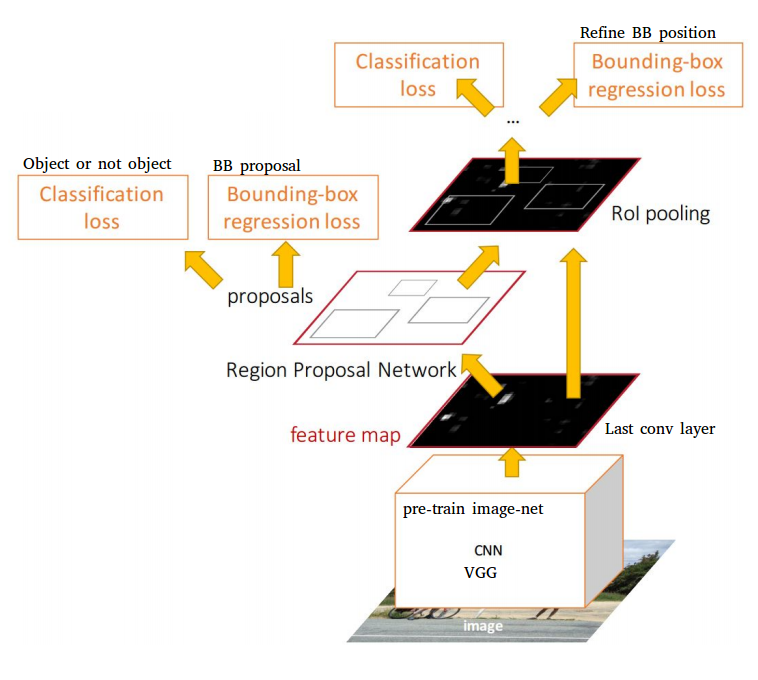}
   \caption{Faster R-CNN scheme. Ren et al [4].}
\end{figure}

\section{\textbf{Methods}}

\subsection{Image pre-processing}
The image set contained hand X-ray  images, especially of the distal radius area. The X-ray images were taken at Hadassah hospital, Jerusalem and analyzed by a hand orthopaedist expert at Hadassa Hospital.
The ray images were taken and processed by GE Healthcare Revolution XR/d Digital Radiographic Imaging system, Agfa ADC Compact digitizer and GE Healthcare thunder platform DIACOM. The  initial dataset pool before augmentation for training and test together contained 55 AP  
images with distal radius fractures and 40 AP images of hands with no fractures. In addition, there were 25 original images not showing hand bones, for  the negative image set. These images were divided into  training and  test sets. About 80\% for training and 20\% for the test.
The initial dataset images of the  positive images were augmented  to 4,476 images with labels and bounding boxes for each augmented image, using mirroring, sharpness, brightness and contrast augmentation. We didn't use shear, strain or spot noise augmentation since these could cause a normal hand image to be classified as a hand with a fracture. The images were annotated by VOTT software.
     
\subsection{CNN based model, research approach.}     
 
We used Faster R-CNN inside Microsoft's cognitive toolkit deep learning framework. The neural network for classification in the Faster R-CNN is VGG 16.  For transfer learning the VGG 16 in the Faster R-CNN we used the image-net database. The X-ray images in the dataset came in couples: anteroposterior position image, AP and lateral position image, of the same hand. After some testing we found that the object detection neural network had better results when  trained only on AP images instead of AP and lateral images together as the AP and lateral images are substantially different. Another reason is in some of the lateral images the fracture does not appear because the other bones hide the fracture. 

Adjusting the hyperparameters such as: nms, roi, and bounding box scale and ratio in the region proposal, regression and evaluation stages  significantly improved, by  a factor of 2 the mAP score and in particular the network's precision of finding the location of the fractures.

To increase the classification accuracy of finding if fractures appear or not in the   X-ray image  the X-ray images were tagged with two labels one for images with fractures and one for hand images with no fractures. This method was used since there is a large similarity between the hand  X-ray images with fractures and without fractures. In this way, the deep neural network find and trains on the differences between the two kinds of images. For the negative images in the training, images of different kinds, not related to X-ray images were added.

To increase the detection accuracy, four types of image augmentation were created: sharpness, brightness,  contrast, and mirror symmetry.

Other ways initially considered in the research are: Training a VGG only, network \cite{simonyan2014very}. The disadvantage, it can only classify if there is a distal radius fracture or not in the image. It does not detect the fracture location in the image. Training an SSD 500 network for object detection \cite{liu2016ssd}. Its disadvantage, the highest input image resolution allowed is 512x512 pixels resulting in lower mAP score and accuracy compared to Faster R-CNN, especially in detecting small objects like bones fractures. 

\section{\textbf{Experiments}}
We ran the Faster R-CNN in Microsoft cognitive toolkit framework on NVIDIA GeForce GTX 1080 Ti graphic card. With 3584 gpu cores, 11.3 TFLOPS, 11 Gbps GDDR5X memory, 11 GB frame buffer and 484 GB/sec memory bandwidth. With CUDA version: 9.0.0 and CUDNN version: 7.0.4, on a computer with an Intel Core i7-4930K processor. Which has six cores each operating at a frequency of 3.4-3.9 GHz, twelve threads and bus speed  5 GT/s DMI2.  The computer was configured to the Debian 9, operating system.

The possibility to train high-resolution images is an important advantage of Faster R-CNN when it used for X-ray images of fractures, as often the fracture is small  compared to the total image.
We tested image sets with different  resolutions from $500\times 500$ to $1600\times 1600$ pixels. We got the best results when the image resolution is $1300 \times 1300$ pixels.


\vspace{5mm}

 
 
 

\begin{table}
\caption{Object detection precision versus number of images augmentation}
\begin{center}
\begin{tabular}{>{\hspace{2pc}}l<{\hspace{20pt}}l<{\hspace{50pt}}l}
 Mean AP & Loss &  Number of augmented training images.\\
 \hline

0.657  & 0.061 & 552 \\
 
0.825 & 0.026 & 4280 \\
 
0.866 & 0.020 & 4476 \\
\hline
\end{tabular}
\end{center}
\end{table}

Training the two label Faster R-CNN  with a set of 4,476 augmented images, resulted in a training loss  of 0.020. Testing this network with 1,312 augmentations of images, which were not included in training part of the data set. Received a mAP of 0.866.  Using mean AP average accuracy as defined in The PASCAL Visual Object Classes challenge, VOC2007 \cite{everingham2010pascal}. 
The classification accuracy for a distal radius fracture is 96\%. 
We ran the system for 45 epochs. It almost converged in the 22nd epoch.

Figure 2 shows hand X-ray images. On the left column are images before evaluating by the deep neural network system for object detection.  On the right columns are the same X-ray images after evaluation. The system detected the distal radius fracture and labeled it with the word fracture in the image. The system located the fracture locations and marked their locations in the image with a blue rectangle. The number in the images is the certainty of the fracture detection, as calculated by the system. The certainty ranges from 0 to 1 where 1 is full certainty.

\begin{figure}[h!]

\includegraphics[width=0.5\linewidth, height=8cm]{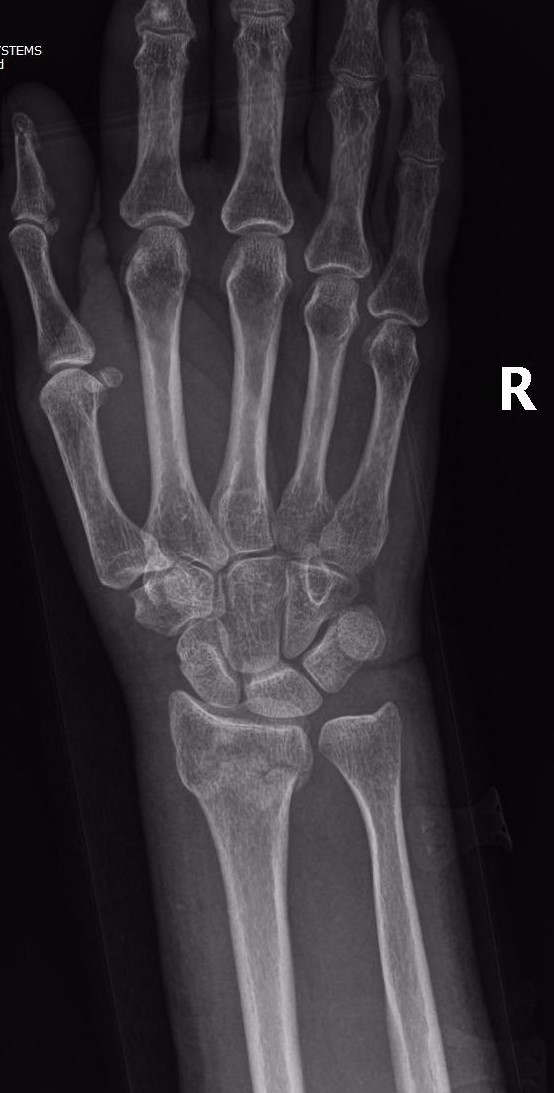} 
\includegraphics[width=0.5\linewidth, height=8cm]{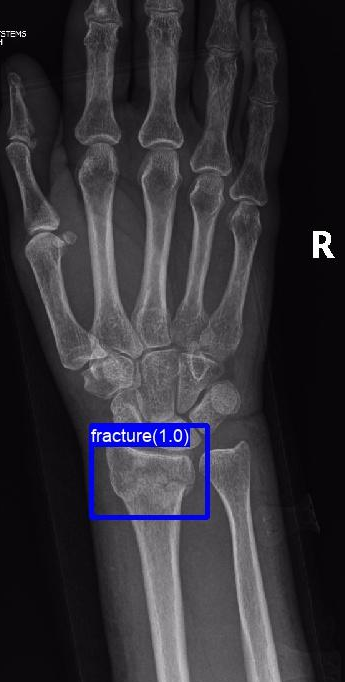}
\includegraphics[width=0.5\linewidth, height=8cm]{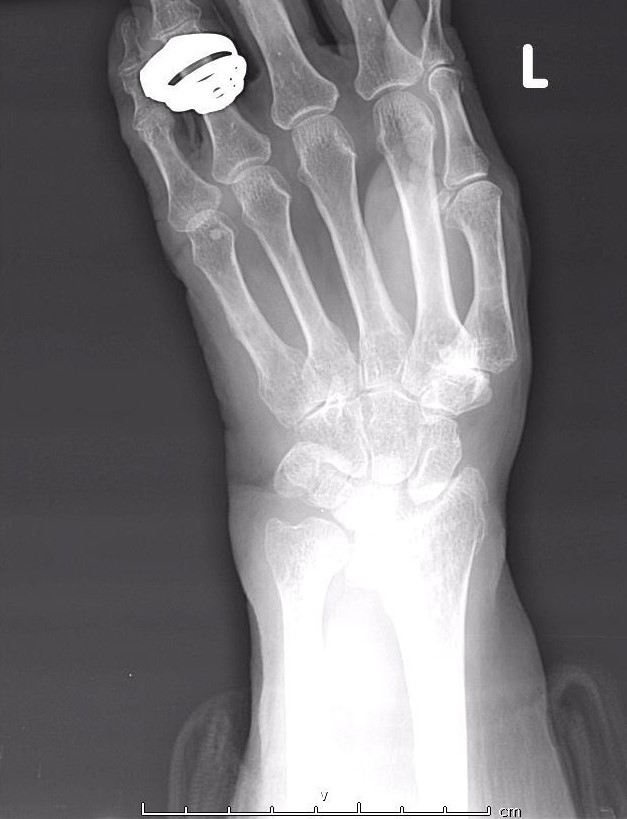} 
 \includegraphics[width=0.5\linewidth, height=8cm]{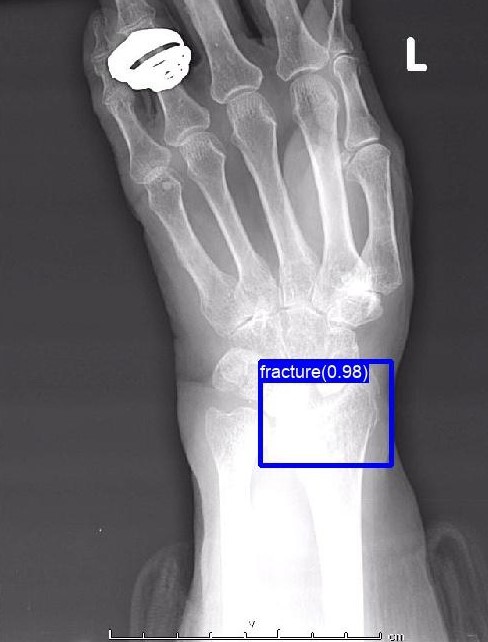}
\caption{X-ray images of hands bones. Before and after evaluation by the machine vision system. The distal radius fractures detected by the deep neural network are bounded by a blue rectangle. }
\label{fig:a}
\end{figure}





In figure 2, the lower  images are over-exposed. It can be seen that even in this case the deep network system was able to detect the fracture with very high certainty. On the other hand, a radiologist or physician would have asked to make another image, as by  regular standards this image cannot be diagnosed\newline

\begin{figure}[h!]

\includegraphics[width=0.5\linewidth, height=10cm]{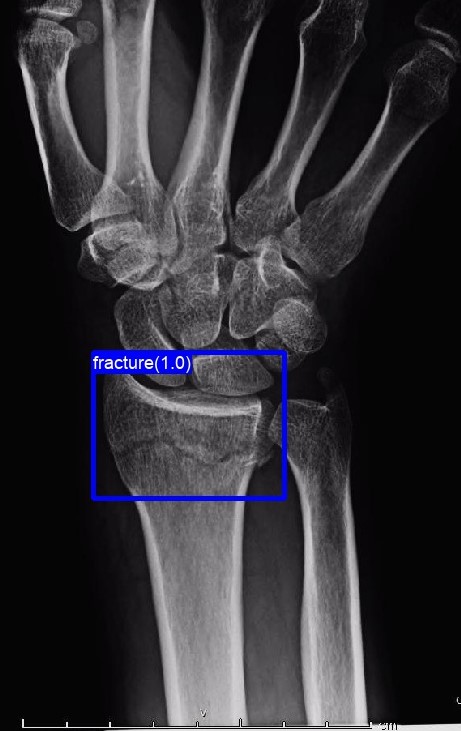} 
\includegraphics[width=0.5\linewidth, height=10cm]{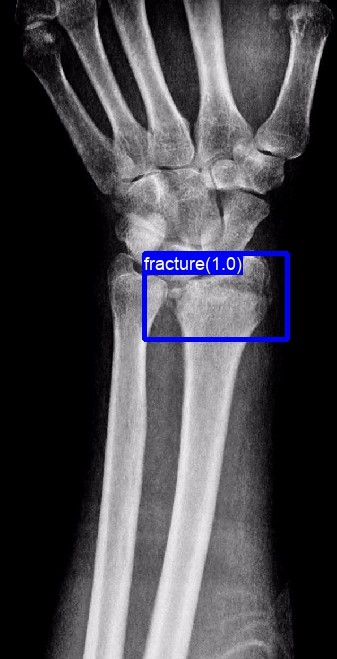}
\caption{Distal radius fracture detection in augmented X-ray images with enhanced contrast.}
\label{fig:b}
\end{figure}

Figure 3, shows detection of fractures by the object detection deep neural network in X-ray images with enhanced contrast, from the augmented images in the test set. 

\section{\textbf{Discussion}}
\subsection{Conclusions}
In this paper, the neural network, Faster R-CNN, was trained to detect distal radius fractures. The system was trained on 4,476 augmented anteroposterior hands  X-ray images and pre-trained on the imagenet data set. Testing this deep neural network system on 1,312 anteroposterior hands X-ray images resulted in an mAP of 0.866 for object detection of distal radius fractures, With an accuracy of 96\% in classifying if there is a distal radius fracture. The network accuracy is substantially higher than the average  accuracy of fracture identification by qualified radiologists and physicians. This makes the Faster R-CNN  an excellent tool to detect hand fractures in X-ray images. We obtained this accuracy using initial small data set of only 38  anteroposterior X-ray images,  of hands diagnosed with fractures. As far as we know it is the smallest set of positively diagnosed images used to train a machine vision network successfully for classification or objects detection, on  X-ray, CT or MRI  medical images.

The ability, we showed in this paper, to train Faster R-CNN with a small number of X-ray images with distal radius fractures and get high accuracy detection may open the possibility for detecting  rare diseases or rare symptoms of common diseases, by Faster R-CNN or related deep neural networks. Where the appearance of these diseases or symptoms is diagnosed today by X-ray, CT or MRI images. But since there is  only a small number of diagnosed X-ray images available, of these medical cases, only experts can detect them. Rare diseases conditions that affect less than 1 in 1800 individuals often go untreated. On average, it takes most rare disease patients eight years to receive an accurate diagnosis, \cite{OMS}
By that time, they have typically seen 10 specialists and have been misdiagnosed three times. The deep neural network  object detection may make the diagnosis of some rare diseases available to every physician. 

\subsection{Limitations and future scope.}
The network was trained on a small set of images compared to the regular training for machine vision detection of X-ray images, which varied from about thousand original images, \cite {kim2018artificial}, up to tens of thousands original images, \cite{dong2017learningrathode2015mri}. For regular images, not X-ray images or medical images, Faster R-CNN already achieved good precision results for labels trained by a small set of  images, \cite {karol}.
In the future testing and training the Faster-RCNN network with thousands of images is suggested to verify the accuracy robustness, of the network and probably to increase the accuracy more. Although the network achieved top-level accuracy, much higher than regular physician or radiologist accuracy. Every single percent increase in the accuracy means many more detections that are correct in a global pool of tens of millions of new X-ray bone fractures imaged every year. 

Training the network for detection of rare diseases facing a challenge of finding positively diagnosed X-ray images. In the future, we suggest collecting positively diagnosed X-ray images of rare diseases from different hospitals and medical centers. The collection can be on a national scale or even international scale. Examples of rare diseases that are diagnosed by X-ray images, Pycnodysostosis, Erdheim-Chester disease, Gorham's disease,  adult-onset Still’s Disease (AOSD), Carcinoid tumors, Pulmonary hypertension and Chilaiditi's syndrome. Once enough X-ray images of rare diseases will be collected to train the objects detection neural network the system could be distributed to any medical facility, substantially aiding  the  detection of  rare diseases and significantly reducing the time till people who have the rare disease get the proper treatment

\bibliographystyle{unsrt}
\bibliography{sample1}

\end{document}